\documentclass[11pt, a4paper]{article}
\usepackage{orcidlink} 
\usepackage[utf8]{inputenc}
\usepackage[T1]{fontenc}
\usepackage[margin=1in]{geometry} 
\usepackage{amsmath}
\usepackage{amssymb}
\usepackage{graphicx}
\usepackage{subfig}
\usepackage{booktabs} 
\usepackage{hyperref} 
\usepackage{authblk}  
\usepackage{subfig}

\usepackage[
  style=authoryear,
  backend=biber,
  natbib=true
  ]{biblatex}

\AtBeginDocument{%
  }

\begin{document}

\title{A Multimodal Transformer Approach for UAV Detection and Aerial Object Recognition Using Radar, Audio, and Video Data}

\author[1]{Mauro Larrat \orcidlink{0009-0008-4963-4625} \\ \texttt{maurolarrat@ufpa.br}}
\author[1]{Claudomiro Sales \orcidlink{0000-0002-2735-1383} \\ \texttt{cssj@ufpa.br}}

\affil[1]{Institute of Exact and Natural Sciences, Federal University of Pará, Brazil}

\date{\today}

\maketitle

\begin{abstract}
    Unmanned aerial vehicle (UAV) detection and aerial object recognition are critical for modern surveillance and security, prompting a need for robust systems that overcome limitations of single-modality approaches. This research addresses these challenges by designing and rigorously evaluating a novel multimodal Transformer model that integrates diverse data streams: radar, visual band video (RGB), infrared (IR) video, and audio. The architecture effectively fuses distinct features from each modality, leveraging the Transformer's self-attention mechanisms to learn comprehensive, complementary, and highly discriminative representations for classification. The model demonstrated exceptional performance on an independent test set, achieving macro-averaged metrics of $0.9812$ accuracy, $0.9873$ recall, $0.9787$ precision, $0.9826$ F1-score, and $0.9954$ specificity. Notably, it exhibited particularly high precision and recall in distinguishing drones from other aerial objects. Furthermore, computational analysis confirmed its efficiency, with $1.09$ GFLOPs, $1.22$ million parameters, and an inference speed of $41.11$ FPS, highlighting its suitability for real-time applications. This study presents a significant advancement in aerial object classification, validating the efficacy of multimodal data fusion via a Transformer architecture for achieving state-of-the-art performance, thereby offering a highly accurate and resilient solution for UAV detection and monitoring in complex airspace.
\end{abstract}

\providecommand{\keywords}[1]{\textbf{\textit{Keywords---}} #1}

\maketitle

\section{Introduction}
The detection and recognition of Unmanned Aerial Vehicles (UAVs), or drones, has emerged as a pivotal component in investigations pertaining to national security and defense, attributed to the considerable proliferation of these apparatuses in both civilian and military contexts \cite{10827393, doi:10.1177/17568293231190017, 9217829, 9765451, 9694151, Utebayeva2023}. Research in this field focuses on developing precise methodologies for identifying, classifying, tracking, and forecasting UAV behavior — tasks that hinge on reliable detection systems. The intricacy involved in the detection of these aerial platforms arises from multiple determinants, particularly their physical attributes—including diminutive dimensions, the reflective properties of materials utilized in their fabrication, operational capacity at low altitudes, biomimetic patterns, stealth technology, and extensive navigational autonomy \cite{10275854, 10028275}. Moreover, environmental factors such as fog, variations in luminosity, and diverse acoustic disturbances, along with both static and dynamic entities within the surveillance domain, can substantively detract from the efficacy of detection frameworks \cite{10274881}. Lastly, technological constraints and financial implications inherent to the implementation of such systems further impose significant limitations on their extensive deployment.

Current research describes various methodologies aimed at mitigating the intrinsic challenges associated with UAV detection and recognition, emphasizing innovations fostered within the domain of Computer Science, particularly through the application of machine learning paradigms supported by diverse data modalities. Nevertheless, the dynamic and complex nature of these challenges necessitates ongoing research and development aligned with technological advancements. Such progress has facilitated the advent of novel multimodal data sources, such as electromagnetic, optical, and acoustic sensors, whose strategic amalgamation can substantially enhance detection efficacy, permitting improved responsiveness to variabilities pertinent to both aerial vehicles and operational contexts. The integration of these disparate data sources, however, engenders additional intricacies pertaining to data heterogeneity and homogeneity—critical considerations in the preprocessing and structuring of information. In this framework, data fusion surfaces as a viable strategy, as it permits the systematic organization and coherent integration of various data types and formats, thus enabling the extraction of more salient features by machine learning algorithms. This approach aspires to elevate the accuracy, robustness, and generalization capabilities of detection systems, thereby contributing to the development of more effective countermeasures against the threats posed by the nefarious utilization of drones.

The escalating deployment of UAVs across civil and military domains introduces novel challenges pertaining to national security, individual privacy, and the safeguarding of critical infrastructure, necessitating the development of more sophisticated and dynamic detection systems. Consequently, this investigation is substantiated by the imperative to enhance the formulation of solutions predicated on contemporary machine learning methodologies, particularly those that utilize multimodal data fusion techniques. These strategies possess the capability to overcome the technical constraints inherent in unimodal approaches, thereby providing increased resilience against environmental variabilities and the interference from extraneous objects. Moreover, the examination of architectures such as the Transformer, which is regarded as one of the most promising frameworks in current research, constitutes a significant scientific advancement in the domain of automated UAV detection and recognition, with practical ramifications across multiple sectors, including public safety, strategic defense, and environmental monitoring.

Building upon the outlined challenges, this study aims to enhance the precision, resilience, and generalizability of UAV detection and recognition in complex environments by integrating heterogeneous data categories pertinent to machine learning \cite{10770236}. To realize this aim, a novel multimodal Transformer architecture is proposed, which innovatively integrates multiple data streams, including audio, visual band video (RGB), infrared (IR) video and radar. This architecture enables the processing of unique features from each independent modality, pooling complementary information within an integrated framework to enable robust discrimination of drone targets from other aerial objects like birds, helicopters, and airplanes, particularly in outdoor conditions. The research further aims to demonstrate how this approach outperforms traditional single-modality systems by improving detection accuracy through effective balancing of sample counts across modalities and addressing modality-specific limitations. Although the class distribution is inherently imbalanced across modalities — with certain classes appearing exclusively in specific sensors — the proposed multimodal Transformer model still achieved excellent performance.

This research adopted an applied approach, primarily quantitative, focusing on the development of solutions for UAV detection and recognition in highly complex scenarios. The study was, in essence, explanatory, investigating the extent to which the fusion of diverse data categories, coupled with the application of an advanced multimodal Transformer architecture, influences the accuracy, robustness, and generalization capacity of detection systems. The technical procedure was experimental, involving the use of two third-party datasets to construct controlled test scenarios. Within these scenarios, the proposed multimodal Transformer algorithm was implemented and rigorously evaluated based on objective performance metrics, demonstrating its capability to process multiple data streams. The methodology addressed class imbalance and modality-specific limitations, with an emphasis on early fusion for data integration. The scientific method employed was primarily deductive, building upon an established model and fusion strategy, with empirical analysis of results guiding adjustments and validations.

Through rigorous empirical evaluations utilizing two distinct third-party datasets \cite{SVANSTROM2021107521, Kang2024}, the model exhibited substantial robustness under various data acquisition conditions, thereby demonstrating its capability for extrapolation to real-world environments. The adoption of an early fusion technique proved computationally advantageous while achieving performance metrics that are either comparable to or exceed those of late fusion methodologies in native multimodal systems \cite{shukor2025scalinglawsnativemultimodal}. Importantly, the approach addresses class imbalance challenges across diverse modalities, thereby ensuring dependable classification outcomes even amidst disparate data distributions. Additionally, the data preprocessing framework, encompassing cyclic reshaping and z-score normalization, preserved data integrity, as indicated by Kullback-Leibler (KL) divergence values remaining proximal to zero across all modalities. These results elucidate the substantial benefits of multimodal data fusion in enhancing detection precision and establishing a robust operational paradigm for complex and noisy environments, thus facilitating prospective advancements in multimodal UAV detection frameworks. This investigation specifically delineates a multimodal Transformer model that amalgamates radar, acoustic, and visual data for UAV detection and aerial object recognition.

\section{Related Research}

The paper \cite{10943055} proposes an intelligent anti-drone surveillance system designed for public security, integrating optical, thermal, RF, and acoustic sensors with advanced machine learning algorithms. The architecture employs late fusion to combine predictions generated by the sensors, synchronizing detection labels, confidence levels, and timestamps. The primary visual detection model is based on YOLO with a CSPDarkNet53 backbone and utilizes PANet for multi-scale detection. Performance metrics reveal high efficacy, with a mAP@0.5 of 0.856, in addition to tracking via DeepSORT and geographic position estimation through camera parameters. Real-world tests conducted with 5G infrastructure demonstrated the superior performance of autonomous networks regarding latency, jitter, and packet loss, highlighting the system's potential in complex urban scenarios. Among the challenges encountered, the need for greater robustness in high-noise environments and the mitigation of false positives are prominent, especially given the evolution of UAV technologies and the presence of stealth drones.

The work by \cite{10018733} proposes a cooperative and autonomous multi-robot system for C-UAS missions, integrating unmanned ground and aerial vehicles with complementary sensors such as LiDAR, radar, stereo camera, and optical cameras. Intruder detection is performed through multimodal perception and data fusion, where LiDAR and the stereo camera identify potential targets, and the camera with radar refines the intruder's position for interception. For data processing, a CNN based on YOLOv4 Tiny is employed, trained on synthetic data and fine-tuned with real images, in addition to depth filtering techniques, 3D reconstruction using a pinhole model, and Kalman filter-based tracking. The coordination of actions between robots is managed by a finite state machine, enabling distributed missions with minimal information sharing. Field tests demonstrated the system's effectiveness but also revealed challenges such as the influence of the intruder's material reflectivity on LiDAR detection, highlighting the importance of sensor fusion to mitigate these limitations.

The paper \cite{10610957} introduces MMAUD, a multimodal dataset that integrates information from stereo vision sensors, multiple LiDARs, radars, and audio arrays, providing a rich and diverse foundation for advanced drone detection techniques. MMAUD's distinctive features include its aerial data collection and the provision of ground-truth data generated by sensors, ensuring high precision and credibility for experiments. The data is organized into six distinct drone categories, including the DJI Mavic2, Mavic3, Avata, Phantom4, M300, and ambient noise sequences. To evaluate the effectiveness of detection algorithms, metrics such as mAP and FPS were utilized, alongside the three-dimensional estimation of drone positions. Among the main challenges encountered were geographical and legal flight restrictions in Singapore, difficulties in sensor synchronization, limitations in the diversity of drone models, and the sampling frequency of the ground truth. Nevertheless, MMAUD represents a significant advance in constructing realistic and comprehensive databases for the development and evaluation of UAV detection and tracking models.

The study \cite{10175389} proposes a method for drone detection and classification through the fusion of data obtained from optical, FMCW RADAR, and acoustic sensors, based on field experiments. The approach is divided into two stages: in the first, individual data from each sensor are processed by pre-trained CNN models (GoogLeNet, ResNet-101, and DenseNet-201) for initial detection and classification; in the second stage, data from all sensors are combined using multinomial logistic regression, aiming to improve overall accuracy. The metrics used include precision, recall, and F-Score, with the fusion of three sensors yielding the best results, particularly with the DenseNet-201 model. One of the main challenges reported was the difficulty in correctly classifying the type of drone, especially due to the similarity between spectra and Doppler signatures, which makes classification less precise than detection. The study demonstrates that sensor fusion is effective for increasing performance in anti-drone surveillance scenarios, provided that individual sensors exhibit minimally reliable performance.

The paper \cite{10229419} presents a machine learning-based approach for UAV detection, utilizing a multimodal dataset comprising images, audio, and RF signals. The authors proposed classification models based on ensemble learning, exploring late fusion of data through soft voting and hard voting applied to classifiers such as CNN and XGBoost. The results demonstrated that the ensemble models surpassed or matched the performance of unimodal models across metrics including accuracy, precision, recall, F1-score, and Type I and Type II error rates, highlighting the robustness of the approach when faced with imprecise or incomplete data. One of the primary challenges encountered in the study was obtaining audio and image samples from online sources, which may have affected variability and, consequently, the models' performance, especially for audio. Nevertheless, the audio-based classifiers showed the best average prediction times, suggesting potential for future optimizations with improvements in preprocessing and dataset expansion.

The paper \cite{9774975} presents an innovative approach based on ensemble deep learning for the detection and classification of unauthorized drones. The proposed model integrates different data categories—acoustic signals, images/videos, and RF signals—by combining synthetic and deep features. The technique employs two binary CNN-based classifiers, integrated through an ensemble learning scheme, aiming to overcome common environmental limitations in classification systems based on isolated features. Data fusion occurs via an ensemble model that exploits the diversity of weak classifiers, leading to a more robust final model. The metrics used for evaluation include accuracy, precision, recall, F1-score, and complementary specificity. Among the challenges faced in the experiment are the specific limitations of each sensor type, such as the recall of cameras to adverse weather conditions and the need for complementary sensors to differentiate UAVs from other aerial objects based on sound or image.

In \cite{10835466}, the study proposes a multimodal approach that combines image, audio, and RF signal data, integrating transfer learning and ensemble stacking techniques for drone detection. Data fusion is achieved by stacking the results from base classifiers—AlexNet (image), CNN14 (audio), and VGG16 (RF)—allowing the weaknesses of each isolated modality to be compensated by others. The best performances were observed in the image and audio classifiers, both achieving 100\% accuracy, precision, recall, and F1-score, while the RF classifier had inferior performance, with 77.34\% accuracy and 73.91\% precision, highlighting the greater variability and complexity of RF signals. The ensemble stacking model achieved 99.22\% accuracy and superior performance in recall (98.39\%) and F1-score (99.19\%), in addition to exhibiting the lowest log loss (0.022) among all models, demonstrating greater robustness. The study also highlights challenges inherent to the variability of RF data and the limited variation in audio and image data, which can influence the results of unimodal classifiers. The proposed approach proves effective in mitigating the limitations of isolated models and reinforces the importance of multimodal fusion for drone detection in complex contexts.

The work in \cite{10640222} proposes a multimodal drone detection framework for static images, utilizing visual (RGB) and thermal infrared (TIR) data, aiming for public security applications. The presented model, named RTM-UAVDet, incorporates a multimodal dynamic convolution (M-DyConv) module and a multi-scale dynamic fusion encoder, enabling efficient extraction of relevant features at different resolutions and modalities without the need for image alignment. Utilizing a robust dataset with over 580,000 annotated images (where regions of interest within the image are identified)—distributed across RGB and TIR modalities—the study demonstrates that the algorithm is capable of handling significant variations in target size, shape, and aspect, as well as challenging scenarios such as low illumination, occlusion, tremors, and partially hidden objects. In experiments, the model achieved an AP\textsubscript{0.5:0.95} of 66.1, the highest among those compared, demonstrating excellent detection capability. Despite not achieving the highest FPS—reaching 72.4 FPS, which is lower than YOLOv5-S (142.3 FPS) and YOLOv6 (135.7 FPS)—RTM-UAVDet presents a good balance between performance and computational efficiency, using only 6.9M parameters and 15.6 GFLOPs, representing a significantly reduced computational cost. These results indicate that, although not the fastest model, RTM-UAVDet offers one of the best precisions with a lightweight architecture, making it suitable for real-time applications in multimodal scenarios.

In the work by \cite{10291920}, the authors propose a drone detection approach that combines infrared and visible images through a multi-scale generative adversarial network (GAN)-based fusion system. Fusion is performed at five different resolutions, which allows for capturing both global and local information and better aligning different image types in the feature space. After fusion, the images are processed by a detection model based on YOLOv5s, which incorporates an enhanced convolutional block attention module (CBAM), responsible for refining spatial and channel information. Experimental results showed significant improvements in precision, recall, and mAP metrics, overcoming limitations of the original YOLOv5 model regarding precision. Despite the advancements, the authors highlight challenges such as the scarcity of specific drone datasets and the absence of tests in an embedded environment with real hardware.

The paper \cite{9415272} presents an algorithm for estimating the position of invisible drones using an acoustic array, employing an A-shaped CNN (a CNN implemented with an up-sampling phase followed by down-sampling). To overcome limitations of optical sensors in situations of occlusion or low illumination, the authors propose converting acoustic data into mel-spectrogram images, enabling the fusion of sound and visual sensors. Fusion is performed based on temporal synchronization between sensors, considering the camera's frame rate (25 FPS) and the audio sampling time (1 second), resulting in 32x40 resolution images per microphone. The A-shaped Neural Network architecture, based on CNN, uses up-sampling techniques to improve the resolution of generated images and down-sampling for 2D coordinate (x, y) regression, estimating the drone's position with a root mean square error (RMSE) of 13.045 pixels. The main challenges addressed include the low detection capability of cameras in dark environments or with obstacles, and the difficulty of extracting acoustic features from overlapping sounds. The use of mel-spectrograms associated with CNNs proved effective in overcoming these limitations by allowing the extraction of acoustic features in a visual format.

While other studies explore interesting directions, many still treat each modality separately or rely on late fusion with specific models. By unifying data streams from multiple sensors at an early stage, our method enables seamless interaction between modalities, leading to a richer environmental perception. This synergy improves UAV detection and recognition while improving system resilience to varying operational conditions and sensor-specific constraints.

\section{Multimodal Datasets for UAV Detection and Recognition}

Multimodal datasets are crucial for UAV detection systems, combining data from various sensors to improve classification accuracy and robustness in real-world conditions. They typically combine measurements from multiple sensing modalities, offering a more comprehensive view of the detection environment. This dynamic integration not only enhances classification performance but also improves real-world robustness.

In this work, we use two third-party datasets, distinct in terms of data acquisition and experimentation \cite{SVANSTROM2021107521, Kang2024}. Our objective is to demonstrate that the model can effectively perform classification regardless of variations in data collection conditions—such as weather, sensor precision, sample quantity, and sensing modalities. By evaluating the model across these heterogeneous datasets, we aim to assess its robustness and potential to generalize to real-world scenarios where multimodal information is gathered under diverse constraints—for instance, along a continental route where an object may be detected in multiple regions using different instruments.

The first dataset obtained from \cite{SVANSTROM2021107521} includes infrared (IR) images, visible-range video, and audio for multi-sensor drone detection, focusing on problems associated with class diversity and sensor fusion. It has recordings on three types of drones: the DJI Hubsan H107D, the DJI Flame Wheel and the DJI Phantom 4 Pro. Other flying objects, such as birds, airplanes, and helicopters, were also recorded to reduce false-positive counts. The dataset includes 650 annotated video clips (365 IR, 285 visible) and 90 ten-second audio recordings featuring drones, helicopters, and background noise, collected from Swedish airports under various weather conditions. Sensor-to-target distances follow the detection, recognition and identification (DRI) standards\cite{sjaardema2015history}, with a maximum of 200 m for drones. A synchronized multi-sensor setup, including IR and visible cameras on a pan-tilt platform, features a fisheye motion-tracking camera, ensuring precise data acquisition. This benchmarked dataset supports the development of UAV detection systems through multi-sensor fusion, with additional technical details available for implementation.

Regarding the first dataset, the audios have been recorded in the Waveform audio file format (WAV), containing signals sampled at 44,100 Hz over two separate audio channels. Each audio file has a constant duration of 10 seconds, resulting in a total of 882,000 samples. The signals are converted into one-dimensional \textit{NumPy} arrays during preprocessing, with each element representing a sample of the audio waveform. This architectural framework facilitates the efficient manipulation of data for ensuing analytical and modeling procedures.

On the other hand, videos are loaded from MPEG-4 Part 14 (MP4) files with an original resolution of 320x256 pixels and a frame rate of 30 frames per second (FPS). To optimize processing, the frames are resized to 128x128 pixels and the frame rate is reduced to 15 FPS. As a result, a video of approximately 10.7 seconds contains 161 frames after reduction. Each frame is stored in a four-dimensional \textit{NumPy} array in the format (number of frames, height, width and color channels), where pixel values are preserved in three channels corresponding to the RGB color components.

We chose this first dataset because adding more entity categories to the data will eventually improve their ability to make that discrimination concerning unmanned aircraft and other airborne entities, by reducing the chances of misidentifying something as a drone when it is actually a bird or a conventional aircraft. This will improve detection accuracy by narrowing down the ambiguity between different classes of objects.

The second dataset consists of radar signals collected only from drone observations \cite{Kang2024}. This dataset does not contain information about the other detectable objects in the first dataset. The radar signals were captured via a 60 GHz radar sensor. The data structure is defined across three key dimensions, essential for sensor operation and environmental interaction. The first dimension consists of 1,000 entries, which refer to unique moments in time, also called frames in time sequence, where the objects in the radar field of view are either detected in motion or appear stationary. On a very high level, these frames are snapshots of scenes recorded during one run. The second part contains 128 subdivisions present in each frame; these windows serve as smaller time windows for further improved temporal resolution. This allows for the observation of the micro-Doppler phenomenon caused by rapid wing flapping or rotor motion, which is crucial for classifying objects by type and activity. The third dimension encompasses 168 functional attributes extracted over each time window. The relationship between multimodal data types, classes, and the number of files is presented in Table \ref{tab:data_classes}.

\begin{table}[htbp]
\caption{Relationship between data types and files for the first \cite{SVANSTROM2021107521} and the second \cite{Kang2024} dataset.}
\label{tab:data_classes}
\begin{minipage}{\columnwidth}
\begin{center}
\begin{tabular}{llcc}
\toprule
\textbf{Data Type} & \textbf{Format} & \textbf{Classes} & \textbf{Number of Files} \\
\hline
Audio & WAV & Background & 30 \\
      &     & Drone      & 30 \\
      &     & Helicopter & 30 \\
\hline
IR Video & MP4 & Airplane   & 76 \\
         &     & Bird       & 79 \\
         &     & Drone      & 157 \\
         &     & Helicopter & 55 \\
\hline
RGB Video & MP4 & Airplane   & 43 \\
          &     & Bird       & 51 \\
          &     & Drone      & 114 \\
          &     & Helicopter & 61 \\
\hline
Radar & Python Pickle (PKL) & Bird  & 18  \\
                       &     & Drone & 36  \\
\bottomrule
\end{tabular}
\end{center}
\bigskip
\footnotesize
\end{minipage}
\end{table}

We selected the second dataset to evaluate how the model handles the imbalance in detection modalities across different classes. In particular, we consider the presence of numerous detectable objects lacking radar modality information as a meaningful challenge for the model, providing a strong basis for assessing its classification performance under incomplete or imbalanced sensing conditions.

These multiple types of data show clear class imbalances, meaning that not all data modalities share the same associated classes across samples. This very fact poses a difficult question in using the multimodal Transformer, whose internal mechanisms must deal with these as part of the classification process. Effectively handling these dimensions is crucial to ensuring robust and accurate performance across all data types.

\section{Data Loading and Preprocessing for Multimodal Aerial Object Detection and Recognition}

\begin{figure*}[!h]
\centering
\includegraphics[width=0.7\linewidth]{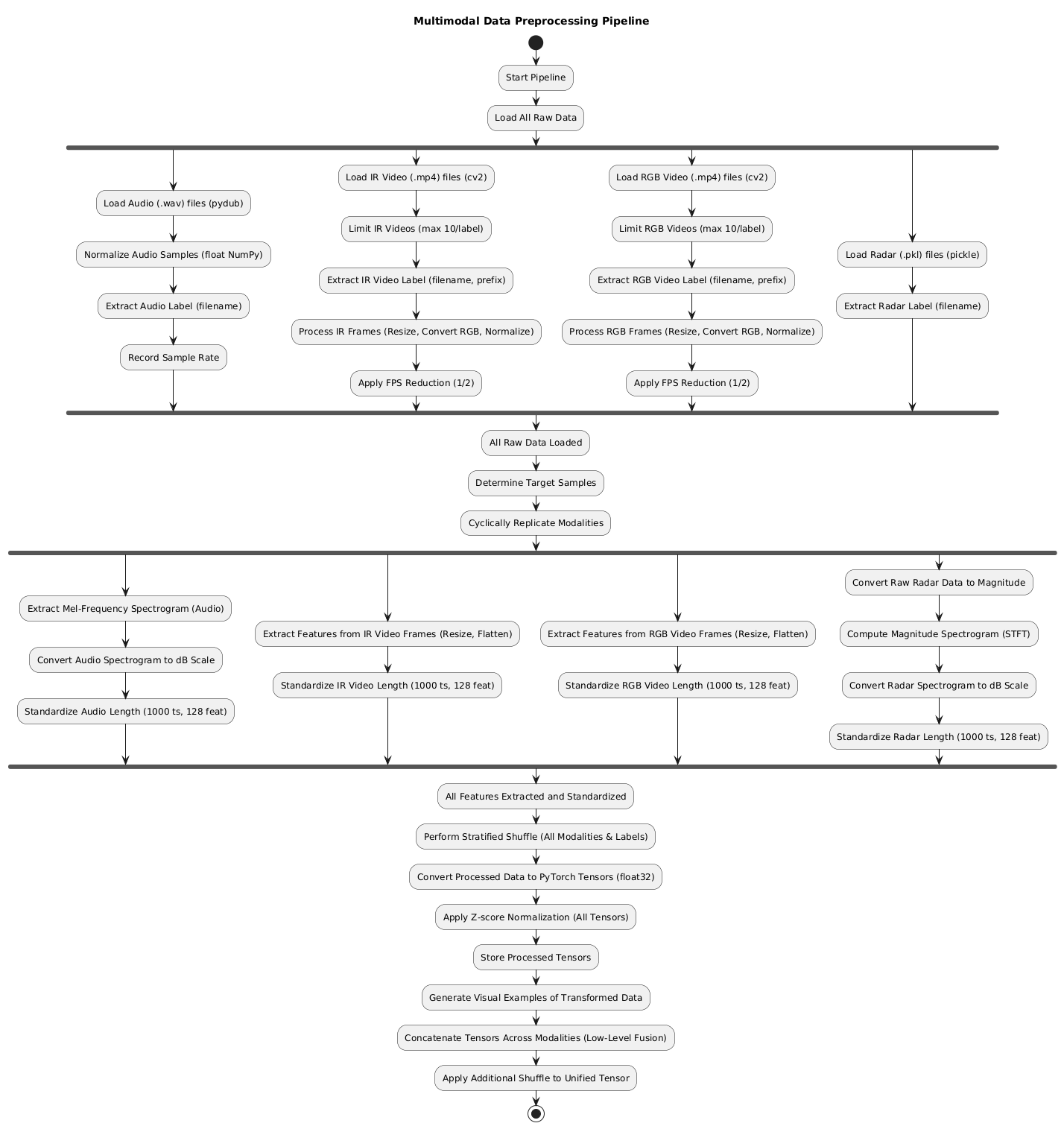}
\caption{This diagram illustrates the sequential and parallel steps involved in preparing data for a multimodal Transformer model. It includes data loading, cyclical replication, feature extraction, standardization, normalization, and final tensor storage.}
\label{DataPreProcessingPipeline}
\end{figure*}

The multimodal data preprocessing pipeline, encompassing audio, video, and radar modalities, is designed to analyze and prepare datasets for training a machine learning model based on multimodal Transformers. The pipeline's general architecture comprises sequential steps: data loading, cyclical replication, feature extraction and standardization, normalization, and tensor storage. The process is illustrated in Figure \ref{DataPreProcessingPipeline}.

The audio module processes recordings in \textit{.wav} format. Each file is associated with its corresponding class, which is extracted from the filename using keywords such as BACKGROUND, DRONE, and HELICOPTER. Python's \textit{pydub} library is utilized to load the audio samples, which are then converted into floating-point \textit{NumPy} arrays, with values normalized according to their sample width (32-bit). The audio samples and their labels are stored in separate lists. The sample rate of the first loaded audio file is recorded and subsequently used for further processing.

\begin{table}[t]
\centering
\caption{Specifications of example files for each modality and the corresponding NumPy array formats.}
\label{tab:data_examples}
\footnotesize
\resizebox{0.7\columnwidth}{!}{%
\begin{tabular}{@{}llcc@{}}
\toprule
\textbf{Modality} & \textbf{Attribute} & \textbf{Example File} & \textbf{Value} \\
\midrule
Audio & Format & BACKGROUND\_001.wav & WAV \\
      & Label &  & background \\
      & Channels &  & 2 \\
      & Sampling Rate &  & 44100 Hz \\
      & Duration &  & 10.00 s \\
      & NumPy Array Shape &  & (882004,) \\
\midrule
IR Video & Format & IR\_AIRPLANE\_001.mp4 & MP4 \\
         & Label &  & airplane \\
         & Original Resolution &  & $320\times256$ \\
         & Reduced Resolution &  & $128\times128$ \\
         & Original FPS &  & 30 \\
         & Reduced FPS &  & 15.00 \\
         & Total Frames &  & 161 \\
         & NumPy Array Shape &  & (161, 128, 128, 3) \\
\midrule
RGB Video & Format & V\_AIRPLANE\_017.mp4 & MP4 \\
          & Label &  & airplane \\
          & Original Resolution &  & $640\times512$ \\
          & Reduced Resolution &  & $128\times128$ \\
          & Original FPS &  & 30 \\
          & Reduced FPS &  & 15.00 \\
          & Total Frames &  & 156 \\
          & NumPy Array Shape &  & (156, 128, 128, 3) \\
\midrule
Radar & Format & mavik\_0\_still.pkl & PKL \\
          & Label &  & drone \\
          & Representation &  & Complex numbers \\
          & Radar Type &  & 60 GHz millimeter wave radar \\
          & $Frames \times Time\,window$ &  & $1000 \times 128$ \\
\midrule
Replication Summary & Audio Samples &  & $40 \rightarrow 200$  \\
                    & RGB Video Samples &  & $40 \rightarrow 200$ \\
                    & IR Video Samples &  & $40 \rightarrow 200$ \\
                    & Radar Samples &  & $58 \rightarrow 200$ \\
\bottomrule
\end{tabular}
}
\end{table}

RGB and IR video files are handled by a dedicated module employing the \textit{cv2} library. This module loads \textit{.mp4} files, and each video's label is determined by its modality prefix (IR or V) and keywords within the filename, such as AIRPLANE, BIRD, DRONE, or HELICOPTER. A mechanism is implemented to limit the number of videos processed per label to 10 due to storage constraints and computational resources. This empirical choice was made as the model showed no significant performance differences with a larger number of videos. For each selected video, frames are extracted, resized to a resolution of 128x128 pixels, and converted to the RGB color space, subsequently normalized to the [0, 1] range. A frame rate reduction is applied, retaining only 1 out of every 2 original frames. The processed frames are then converted into \textit{NumPy} arrays and appended to a list, along with their corresponding labels. The 128-pixel resolution for the spatial dimension of video frames was selected to ensure compatibility with the feature dimension of radar data, facilitating a unified representation within the Transformer's feature space.

Following the loading phase, the raw data undergo a dedicated preprocessing step that starts with the cyclical replication of video (RGB and IR) and radar modalities, as detailed in Table~\ref{tab:data_examples}. This procedure ensures that the sample counts of all modalities match that of the audio modality, which serves as the reference for the overall dataset size.

\begin{figure}[!h]
  \centering

  \begin{minipage}{0.45\linewidth}
    \centering
    \subfloat[Audio.]{\includegraphics[width=\linewidth]{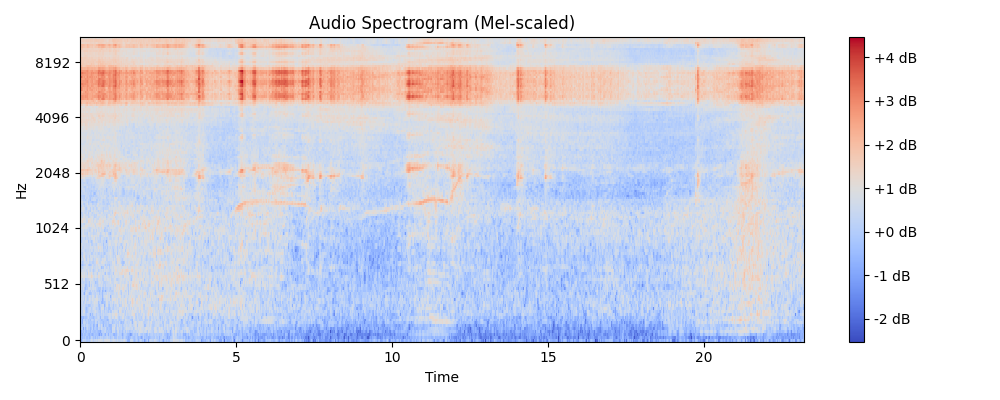}}
    
    \vspace{0.2cm}

    \subfloat[Raw frame from RGB.]{\includegraphics[width=0.67\linewidth]{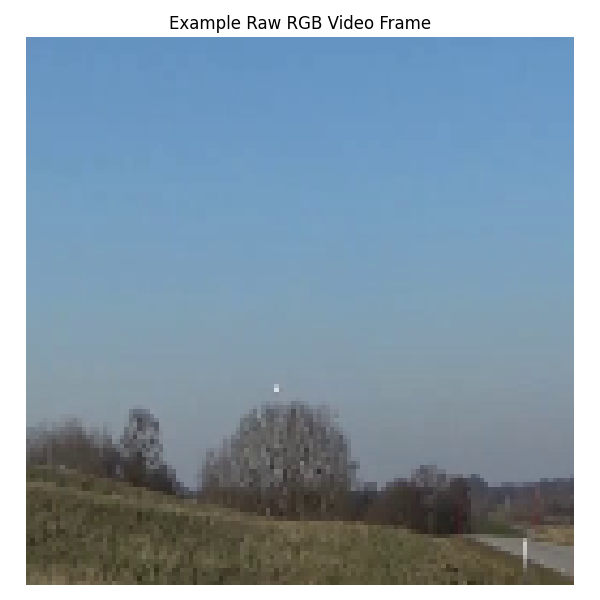}}
  \end{minipage}
  \hfill
  \begin{minipage}{0.45\linewidth}
    \centering
    \subfloat[Radar.]{\includegraphics[width=\linewidth]{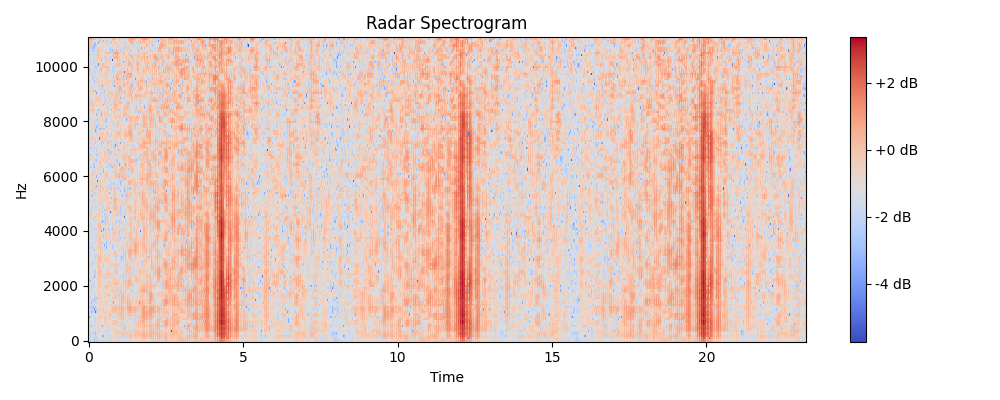}}
    
    \vspace{0.2cm}

    \subfloat[Raw frame from Infrared.]{\includegraphics[width=0.67\linewidth]{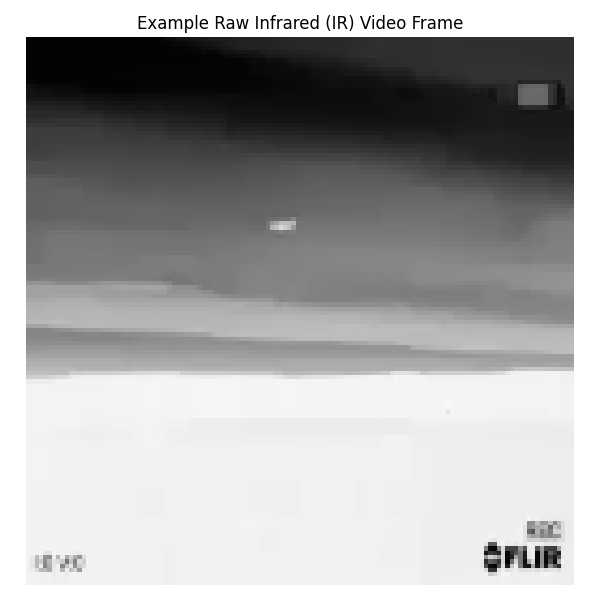}}
  \end{minipage}

  \caption{Audio and radar spectrograms, and raw frames from RGB and IR video modalities.}
  \label{fig:spectrograms}
\end{figure}

Subsequently, features are extracted and standardized for each modality (Figure \ref{fig:spectrograms}):
\begin{itemize}
\item Audio: Audio samples are converted into Mel-frequency spectrograms and then transformed to the decibel scale. The feature sequence is then standardized to a length of 1000 time steps and a dimension of 128 features.
\item Video (IR and RGB): For each video frame, features are extracted by resizing to a dimension of 128 and flattening, resulting in a sequence of feature vectors. The video feature sequences are then standardized to 1000 time steps and 128 features.
\item Radar: Raw radar data, which can be complex, are converted to their magnitude and then to magnitude spectrograms using the Short-Time Fourier Transform (STFT). The spectrograms are converted to the decibel scale and finally standardized to a length of 1000 time steps and a dimension of 128 features. 
\end{itemize}

The choice of 1000 time steps matches the maximum temporal length in the radar data, ensuring consistent temporal representation across modalities. After preprocessing, feature data and labels are stratified and shuffled together to preserve their association and maintain class distribution.

The processed data for each modality (audio, IR video, RGB video, and radar) are converted into \textit{PyTorch} tensors of \textit{float32} type. Subsequently, these tensors undergo z-score normalization, where each tensor's mean is adjusted to zero and its standard deviation to one, across both temporal and feature dimensions. This normalization is crucial for stabilizing the training process.

All data tensors and their corresponding labels were successfully loaded and concatenated, resulting in a combined tensor with shape \textit{torch.Size([800, 1000, 128])}. This concatenation acts as a form of low-level fusion, merging heterogeneous modality-specific features into a unified representation. As a result, the output tensor becomes modality-agnostic — a homogeneous structure that abstracts away the origin of each feature vector. This unified tensor is the input that will ultimately be used by the multimodal Transformer model.

After concatenation, the dataset was shuffled once again to ensure randomness before splitting. These comprehensive data preparation steps — from loading, cyclic replication, early-level fusion and shuffling to storage — were completed in approximately 15 minutes on a system equipped with an AMD Ryzen 5 8600G w/ Radeon 760M Graphics (4.35 GHz) processor and 16.0 GB of RAM (15.1 GB usable), running Windows 11 Pro (64-bit operating system, x64-based processor). 

\subsection{Qualitative Validation of Data Cyclic Replication}

\begin{figure}[!h]
    \centering

    \begin{minipage}{0.48\linewidth}
        \centering
        \subfloat[Visual Band Video (RGB). \label{subfig:rgb}]{
            \includegraphics[width=\linewidth]{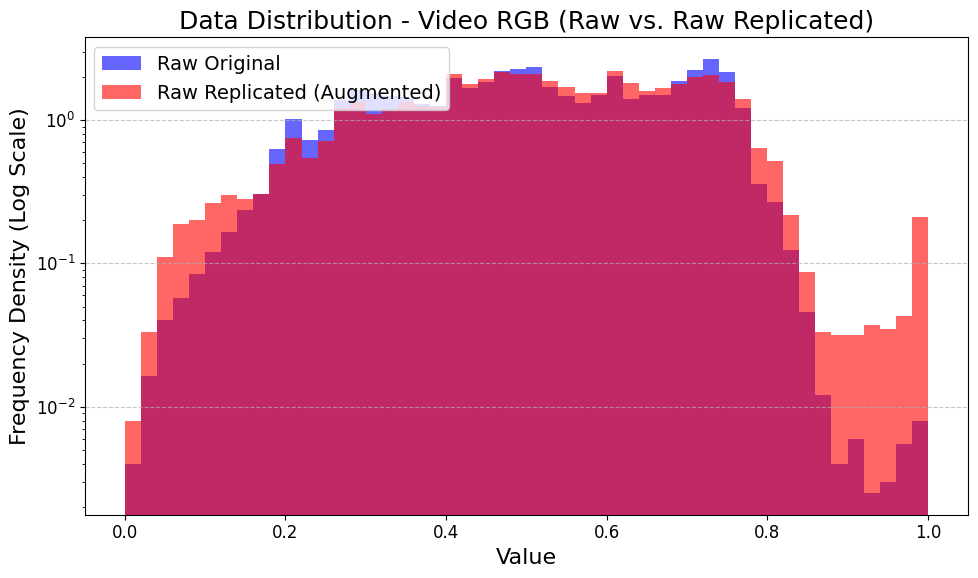}
        }
    \end{minipage}
    \hfill
    \begin{minipage}{0.48\linewidth}
        \centering
        \subfloat[Infrared (IR). \label{subfig:ir}]{
            \includegraphics[width=\linewidth]{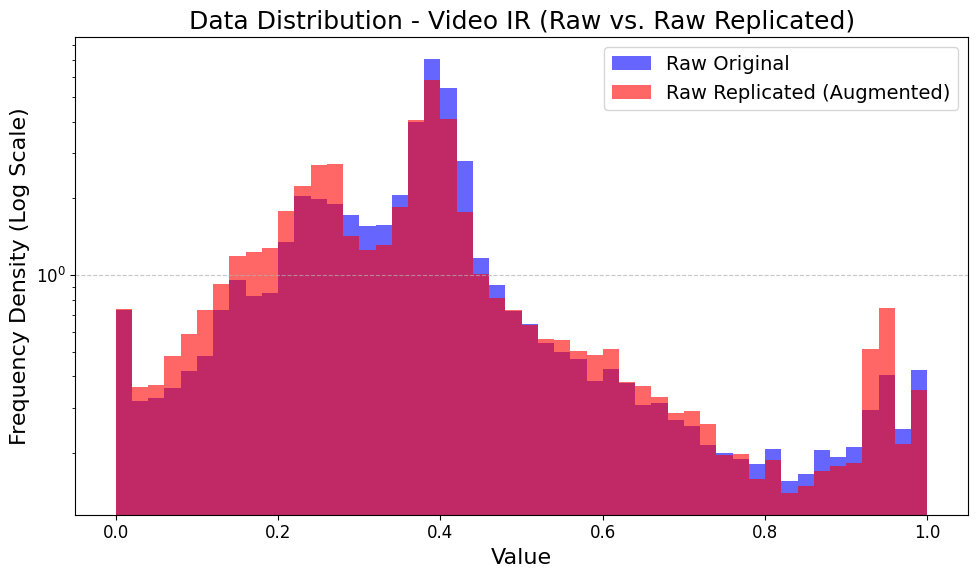}
        }
    \end{minipage}

    \vspace{0.5cm}

    \begin{minipage}{0.48\linewidth}
        \centering
        \subfloat[Radar Signal. \label{subfig:radar}]{
            \includegraphics[width=\linewidth]{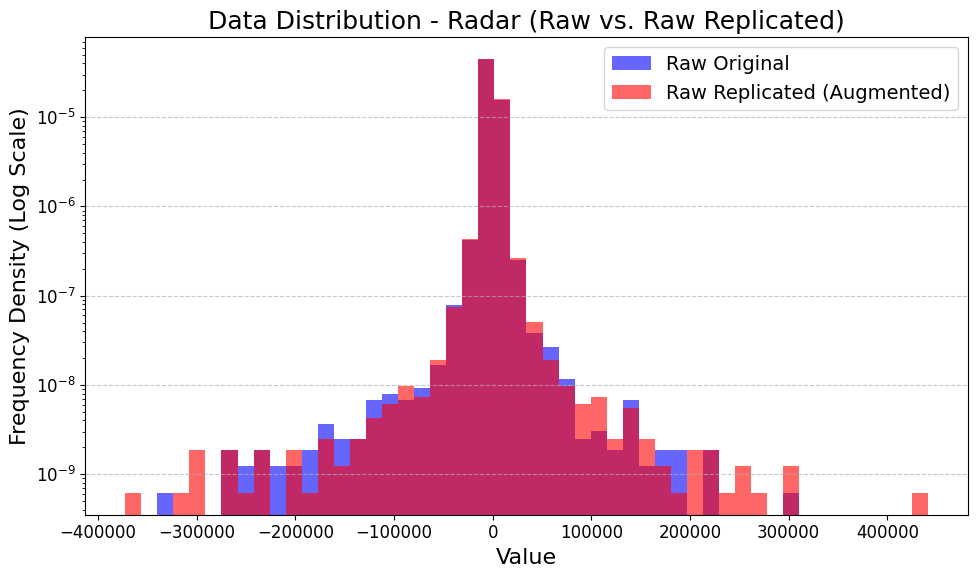}
        }
    \end{minipage}
    \hfill
    \begin{minipage}{0.48\linewidth}
        \centering
        \subfloat[Acoustic. \label{subfig:acoustic}]{
            \includegraphics[width=\linewidth]{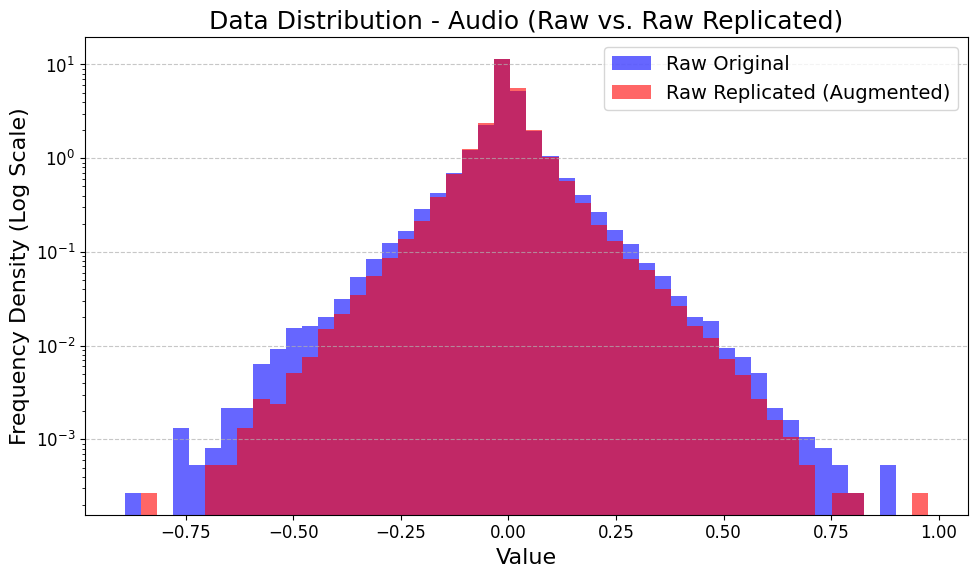}
        }
    \end{minipage}

    \caption{Data distribution comparison between Raw Original and Raw Replicated (Augmented) for various modalities.}
    \label{DataDistBefAfter}
\end{figure}

To ensure consistent sample counts across modalities, audio, RGB video, IR video, and radar samples underwent cyclic replication to match a standardized count of 200 samples. This process was critical for harmonizing data dimensions, enabling seamless integration into the subsequent machine learning framework. As demonstrated in Figures \ref{DataDistBefAfter}(a)-(d), cyclic replication effectively increased the number of data points without altering their original statistical distributions.

A comparative analysis of the Raw Original and Raw Replicated (Augmented) histograms in Figures \ref{DataDistBefAfter}(a)-(d) provides qualitative evidence that cyclic replication successfully increased the effective sample size for each modality while preserving the intrinsic data characteristics.

For Audio data (Figure \ref{DataDistBefAfter}(d)), the Raw Original distribution, spanning values from -1.0000 to 1.0000, exhibited a mean of -0.0001 and a standard deviation of 0.0863 across 26,460,092 points. The Raw Replicated (Augmented) data, with 26,460,090 points, maintained a highly similar distribution, showing a mean of -0.0001 and a standard deviation of 0.0844. Both distributions are visibly centered around zero with a symmetric spread.

Regarding Video IR data (Figure \ref{DataDistBefAfter}(b)), the Raw Original dataset (228,507,648 points) ranged from 0.0000 to 1.0000, with a mean of 0.3679 and a standard deviation of 0.1859. The Raw Replicated (Augmented) data (229,195,776 points) closely mirrored this, with a mean of 0.3806 and a standard deviation of 0.1828. The histograms show a prominent peak around 0.3-0.4, with a secondary peak near 1.0, and the replicated data accurately reflects these features.

The Video RGB data (Figure \ref{DataDistBefAfter}(a)) also demonstrated consistency. The Raw Original distribution (228,261,888 points), with values from 0.0000 to 1.0000, had a mean of 0.5052 and a standard deviation of 0.1740. The Raw Replicated (Augmented) data (227,917,824 points) maintained a similar profile, with a mean of 0.5087 and a standard deviation of 0.1936. The distributions show a broad, relatively flat peak between approximately 0.2 and 0.8, with some density towards 1.0, which is well-preserved.

Finally, for Radar data (Figure \ref{DataDistBefAfter}(c)), both the Raw Original and Raw Replicated (Augmented) datasets consisted of 645,120,000 points. The Raw Original had a mean of 16.8991+14.7315j and a standard deviation of 10731.9004, while the 'Raw Replicated (Augmented) showed a mean of 16.5455+14.8850j and a standard deviation of 10407.6533. The histograms reveal a strong central peak near zero, indicating a high concentration of values around the origin, with tails extending to larger magnitudes, a characteristic consistently observed in both the original and replicated data.

\begin{figure}[!h]
    \centering

    \begin{minipage}{0.48\linewidth}
        \centering
        \subfloat[Visual Band Video (RGB). \label{subfig:rgb_processed}]{
            \includegraphics[width=\linewidth]{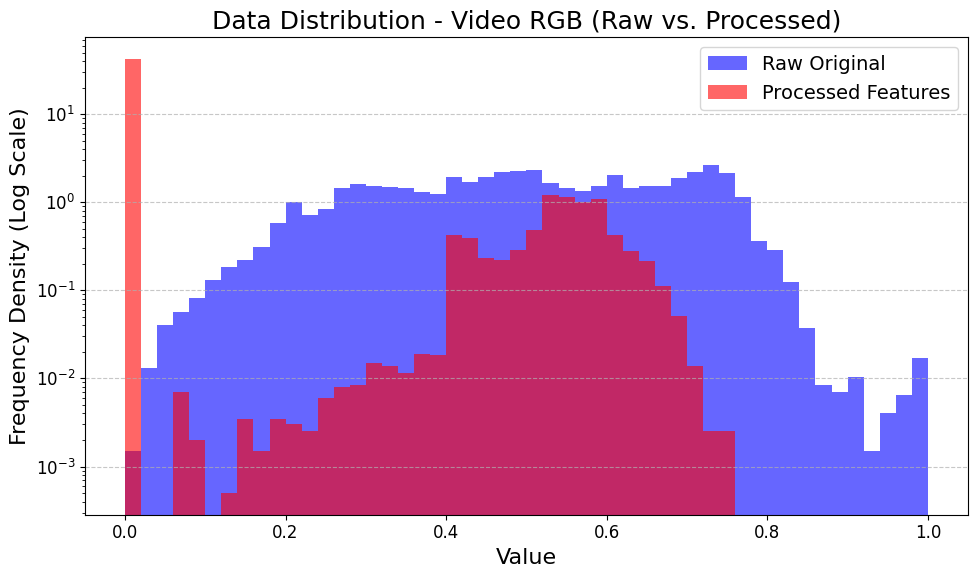}
        }
    \end{minipage}
    \hfill
    \begin{minipage}{0.48\linewidth}
        \centering
        \subfloat[Infrared (IR). \label{subfig:ir_processed}]{
            \includegraphics[width=\linewidth]{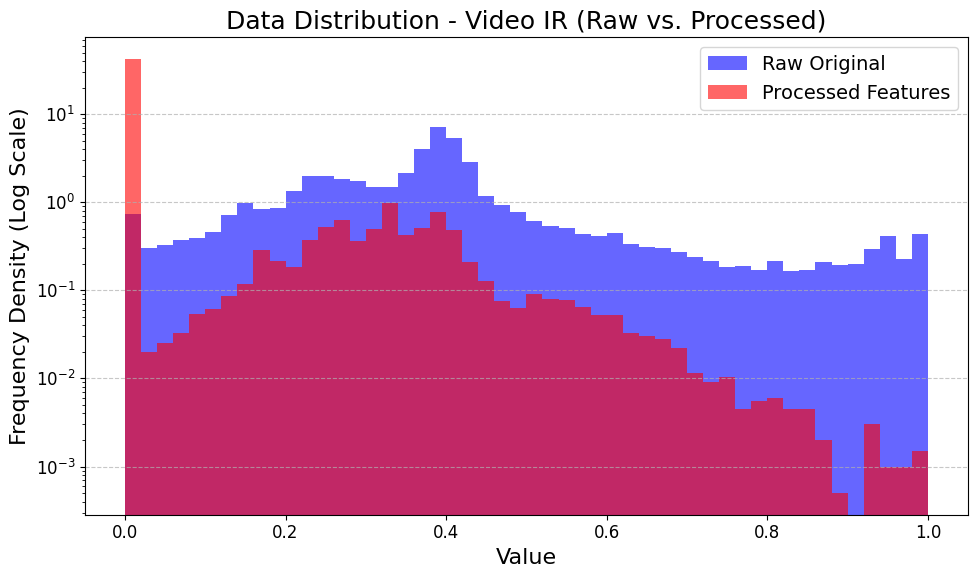}
        }
    \end{minipage}

    \vspace{0.5cm}

    \begin{minipage}{0.48\linewidth}
        \centering
        \subfloat[Radar Signal. \label{subfig:radar_processed}]{
            \includegraphics[width=\linewidth]{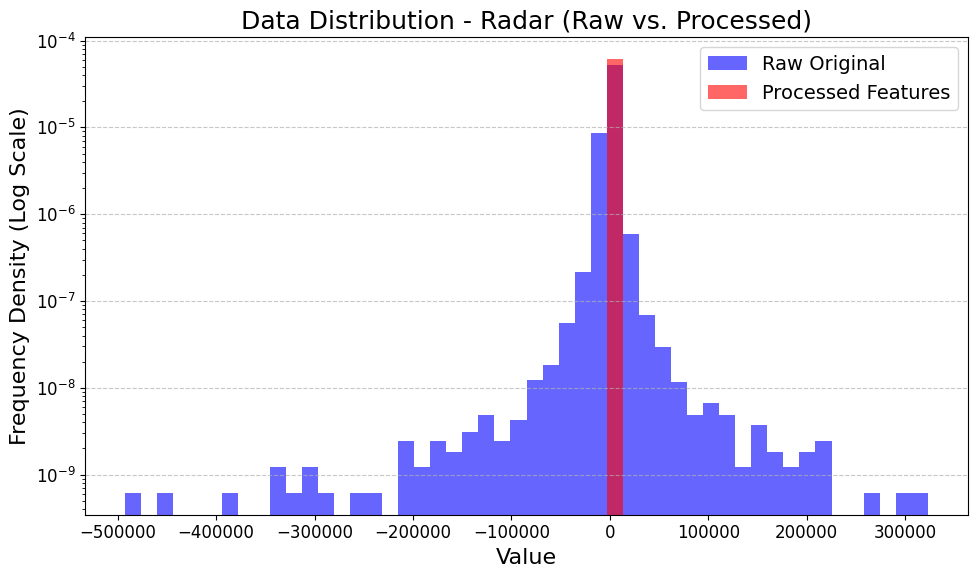}
        }
    \end{minipage}
    \hfill
    \begin{minipage}{0.48\linewidth}
        \centering
        \subfloat[Acoustic. \label{subfig:acoustic_processed}]{
            \includegraphics[width=\linewidth]{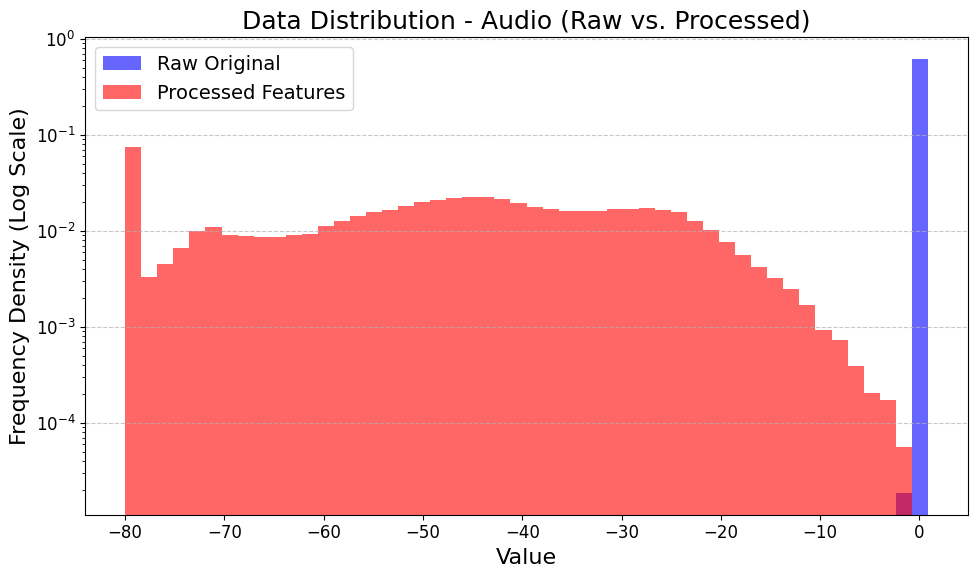}
        }
    \end{minipage}

    \caption{Data distribution comparison between Raw Original and Processed Features for various modalities.}
    \label{DataDistRawProcessed}
\end{figure}

The consistency of distribution shapes after replication—visually and statistically confirms that sample balancing across modalities preserves the data’s inherent characteristics, which is essential for reliable model training. This integrity is further maintained through the transformation of raw sensor data into lower-dimensional, semantically richer features, as shown in Figure~\ref{DataDistRawProcessed}(a)--(d).

For Radar data (Figure \ref{DataDistRawProcessed}(c)), the transformation from Raw Original to Processed Features is substantial. The raw data, consisting of 645,120,000 points with complex values (mean 16.8991+14.7315j, standard deviation 10731.9004), exhibited a broad distribution centered near zero. The Processed Features, however, comprises only 3,840,000 points and is characterized by real values ranging from -80.0000 to -0.0390, with a mean of -36.4307 and a standard deviation of 9.2168. This shift indicates a conversion to a spectral representation, which compresses the dynamic range and highlights frequency-domain characteristics more relevant for pattern recognition. The dramatic reduction in data points (from 645M to 3.84M) signifies effective dimensionality reduction, focusing on salient features of the radar signal.

In the case of Video IR data (Figure \ref{DataDistRawProcessed}(b)), the Raw Original (228,507,648 points, range 0.0000-1.0000, mean 0.3679, std 0.1859) showed a distribution with a prominent peak around 0.3-0.4. The Processed Features (3,840,000 points) retained the 0.0000-1.0000 range but underwent a significant shift in its distribution, with a mean of 0.0491 and a standard deviation of 0.1265. The histogram of processed features is heavily skewed towards zero, suggesting that the feature extraction process might be emphasizing low-intensity thermal information or performing a non-linear transformation that concentrates pixel values at the lower end of the spectrum, potentially highlighting specific thermal signatures crucial for object detection.

Similarly, Video RGB data (Figure \ref{DataDistRawProcessed}(a)) exhibited a clear transformation. The Raw Original (228,261,888 points, range 0.0000-1.0000, mean 0.5052, std 0.1740) had a relatively uniform distribution across its range. The Processed Features (3,840,000 points), while maintaining a range from 0.0000 to 0.9850, showed a mean of 0.0847 and a standard deviation of 0.1996. The processed histogram also displays a strong concentration near zero, indicating that the visual features extracted are likely emphasizing specific image characteristics such as edges, textures, or low-level visual cues, which are more pertinent for a learning model than raw pixel intensities. The increased standard deviation compared to the raw data, despite the shift in mean, suggests a spread in these extracted features.

For Audio data (Figure \ref{DataDistRawProcessed}(d)), the transformation from raw samples to Processed Features is highly distinct. The Raw Original (26,460,092 points, range -1.0000-1.0000, mean -0.0001, std 0.0863) displayed a typical symmetric distribution around zero. The Processed Features (3,840,000 points) occupy a drastically different range, from -80.0000 to 0.0000, with a mean of -48.7434 and a standard deviation of 18.0200. This is characteristic of Mel spectrogram features, where magnitudes are typically converted to a logarithmic (decibel) scale, resulting in negative values. This transformation extracts frequency content over time, which is fundamental for speech recognition, environmental sound classification, or other audio-based tasks. The shift in mean and increased standard deviation reflect the emphasis on spectral patterns rather than amplitude.

In summary, the feature extraction process effectively transforms the diverse raw sensor data into modality-specific representations that are more compact, informative, and mathematically amenable for machine learning. The varying ranges, means, and standard deviations among the processed features reflect their distinct nature and the specialized transformations applied. These processed feature sets, though numerically distinct, are now in a suitable format for concatenation as input to a multimodal Transformer model. This enables the Transformer's self-attention mechanisms to effectively learn intricate intra-modal relationships within each feature set, as well as complex cross-modal dependencies by comparing and integrating information across audio, visual (RGB and IR), and radar domains. The model is expected to leverage these harmonized, rich feature representations to achieve robust and comprehensive understanding for downstream tasks by exploiting the complementary information provided by each sensor.

\subsection{Quantitative Validation of Data Cyclic Replication}

To quantitatively assess the preservation of data distributions after cyclic replication, the Kullback-Leibler (KL) Divergence was computed. KL Divergence measures how one probability distribution differs from another, with a value of zero indicating identical distributions. The KL Divergence is defined by Equation \ref{eq:kl_divergence}:

\begin{equation}
 D_{\text{KL}}(P \parallel Q) = \sum_{x} P(x) \log \frac{P(x)}{Q(x)},
\end{equation}
\label{eq:kl_divergence}

\noindent where $$( D_{\text{KL}}(P \parallel Q) )$$ quantifies the divergence between the original distribution P and the distribution Q after cyclic replication. P(x) and Q(x) are the probability mass functions of P and Q respectively, summed over all possible values of x.

The KL Divergence was first calculated to ascertain the impact of the cyclic replication process on the original data distributions. The results are as follows:
\begin{itemize}
    \item Audio (Raw vs. Raw Replicated): $D_{\text{KL}} = 0.0174$
    \item Video IR (Raw vs. Raw Replicated): $D_{\text{KL}} = 0.0351$
    \item Video RGB (Raw vs. Raw Replicated): $D_{\text{KL}} = 0.0440$
    \item Radar (Raw vs. Raw Replicated): $D_{\text{KL}} = 0.0000$
\end{itemize}

These low KL Divergence values for all modalities---especially the exact zero divergence for Radar, and values very close to zero for Audio, Video IR, and Video RGB---provide strong quantitative validation for the cyclic replication methodology. This indicates that the process successfully increased the number of samples for each modality without introducing significant statistical distortions or alterations to their underlying probability distributions. Such preservation of original data characteristics is crucial for maintaining data integrity, ensuring that the augmented dataset remains a faithful representation of the real-world phenomena and is suitable for subsequent model training without bias from the replication.

Subsequently, KL Divergence was computed to evaluate the distributional changes incurred during the feature extraction process, where raw data is transformed into more compact and informative features. The results are presented below:
\begin{itemize}
    \item Audio (Raw vs. Processed): $D_{\text{KL}} = 11.2957$
    \item Video IR (Raw vs. Processed): $D_{\text{KL}} = 3.0393$
    \item Video RGB (Raw vs. Processed): $D_{\text{KL}} = 5.1509$
    \item Radar (Raw vs. Processed): $D_{\text{KL}} = 2.2214$
\end{itemize}

The significantly higher KL Divergence values observed when comparing Raw Original data with Processed Features are an expected and desirable outcome. These large divergences quantify the substantial transformation that occurs during feature engineering. For instance, audio data is converted to Mel spectrograms, and radar data to spectrograms, fundamentally altering their statistical properties from raw time-domain signals. Similarly, video processing extracts specific features that condense information, leading to different distributions.

Specifically, the high divergence for Audio ($11.2957$) highlights the profound change in data representation, as these transformations convert raw sensor readings into abstract, yet highly discriminative, feature spaces. While the raw data captures the physical signal, the processed features are engineered to capture the most salient information for machine learning tasks, often by compressing irrelevant variations and emphasizing patterns. This divergence confirms that the feature extraction process effectively reduces dimensionality and extracts higher-level representations tailored for model learning. These harmonized and information-rich processed features are now prepared for concatenation and subsequent input into a multimodal Transformer model, allowing the model to leverage the complementary strengths of each modality for robust prediction and understanding.

\subsection{Data Normalization and Uniformization}

Following cyclic replication and feature extraction, data from all modalities (audio, RGB video, IR video, and radar) underwent normalization to standardize their scales, a critical step for optimizing deep learning model performance. Given the varying value ranges across sensor types, Z-score normalization was applied, as defined by Equation \ref{eq:z_score}:

\begin{equation}
X' = \frac{X - \mu}{\sigma}
\end{equation}
\label{eq:z_score}

This method transforms the data to have a mean of 0 and a standard deviation of 1, thereby preserving the intrinsic structure of each modality's distribution while ensuring numerical stability for model training.

Subsequently, the normalized data from all modalities were converted into a standardized three-dimensional tensor representation of shape (200, 1000, 128). This uniform framework systematically arranges samples, time steps, and features, respectively, facilitating interoperability and concurrent processing in deep learning architectures. The first dimension (200) corresponds to the standardized sample count for each modality after cyclic replication, which balanced the original counts of 40 for IR and RGB video and 58 for radar to match the 90 samples of the audio modality. The subsequent dimensions, 1000 instants and 128 features per frame, were specifically chosen to align with the sampling resolution of the radar and video data. Finally, a stratified permutation (shuffle) was applied to randomize the order of data samples and their corresponding labels, maintaining their inherent relationships and preserving the integrity of the labeled dataset.

\section{Multimodal Transformer Architecture and Experimental Setup}

This section presents the detailed architecture of the proposed multimodal Transformer model for aerial object detection and recognition, along with the experimental setup including dataset partitioning and training configurations.

\subsection{Multimodal Transformer Architecture}

\begin{figure}[!h]
\centering
  \includegraphics[width=0.5\linewidth]{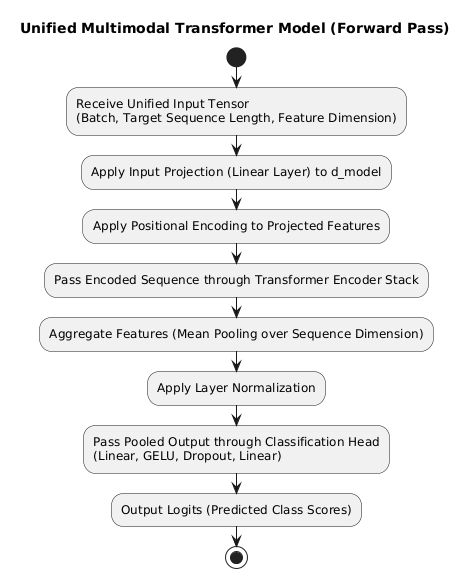}
  \caption{This diagram illustrates the forward pass of the multimodal Transformer, detailing how audio, IR video, RGB video, and radar inputs are processed through individual projections, positional encoding, a shared Transformer encoder, and a final classification head to predict the target class.}
\label{fig:multimodalTrnasformer}
\end{figure}

The proposed multimodal Transformer architecture is specifically designed to process and integrate features from diverse sensory modalities. It operates on an already unified input tensor, formed by concatenating the pre-processed features from audio, infrared video, RGB video, and radar modalities. 

This combined input tensor has a shape of (800,1000,128), where 128 represents the aggregated feature dimension from all modalities at each time step, and 1000 is the standardized sequence length. The architecture consists of an input projection layer, positional encoding, a Transformer Encoder (comprising multiple layers with multi-head self-attention and feed-forward networks), a feature aggregation mechanism, a normalization layer, and a final classifier head, all detailed in the following breakdown of key components.

The architecture, illustrated in Figure \ref{fig:multimodalTrnasformer}, comprises the following key components:
\begin{itemize}
    \item \textbf{Input Projection Layer}: A single shared linear projection layer, \textit{self.input\_projection}, maps the input features from feature\_dim to a shared model dimension, $d_{\text{model}} = 256$. This layer prepares the concatenated multimodal features for the Transformer encoder.
    \item \textbf{Positional Encoding}: Following the input projection, a sinusoidal positional encoding layer, \textit{PositionalEncoding}, is applied. This module injects information about the relative or absolute position of the tokens in the sequence, which is crucial for sequence understanding in Transformer models as they inherently lack recurrence or convolution. The \textit{max\_len} for positional encoding is set to \textit{target\_seq\_len} (1000).
    \item \textbf{Transformer Encoder}: The core of the model is a Transformer Encoder, composed of $num\_layers=2$ identical \textit{nn.TransformerEncoderLayer} modules. Each encoder layer features:
    \begin{itemize}
        \item A multi-head self-attention mechanism with $num\_heads=4$ attention heads, enabling the model to jointly attend to information from different representation subspaces at different positions.
        \item A feed-forward network with an inner dimension of $dim\_feedforward=1024$.
        \item A Gaussian Error Linear Unit (GELU) activation function.
        \item A dropout rate of $0.2$ applied for regularization.
        \item The \textit{batch\_first=True} setting ensures that input tensors are in the format $( \text{batch\_size}, \text{sequence\_length}, \text{feature\_dimension} )$.
    \end{itemize}
    \item \textbf{Feature Aggregation}: After processing through the Transformer Encoder, the output \textit{encoder\_output} is a sequence of contextualized embeddings. These embeddings are aggregated into a single fixed-size representation per sample by applying mean pooling over the sequence dimension (i.e., \textit{encoder\_output.mean(dim=1)}).
    \item \textbf{Normalization Layer}: A \textit{nn.LayerNorm} is applied to the pooled output to stabilize activations and improve training dynamics before the final classification head.
    \item \textbf{Classifier Head}: The final classification is performed by a sequential neural network. This head consists of:
    \begin{itemize}
        \item A linear layer mapping from $d_{\text{model}}$ to $512$ dimensions.
        \item A GELU activation function.
        \item A dropout layer with a rate of $0.2$.
        \item A final linear layer mapping from $512$ dimensions to \textit{num\_classes} (the number of target output classes).
    \end{itemize}
\end{itemize}

\subsection{Experimental Setup}

The dataset was stratified split into training, validation, and test sets with proportions of approximately 60\%, 20\%, and 20\%, respectively, to ensure representative class distribution across all partitions. The detailed class distribution for each set is provided in Table \ref{tab:class_distribution}.

\begin{table}[htbp]
\caption{Class distribution across training, validation, and test sets.}
\label{tab:class_distribution}
\centering
\resizebox{0.7\columnwidth}{!}{%
\begin{tabular}{llccc}
\toprule
\textbf{Set} & \textbf{Class} & \textbf{ID} & \textbf{Samples} & \textbf{Percentage} \\
\hline
\textbf{Training Set (Total: 440)} & airplane & 0 & 55 & 12.50\% \\
                                   & background & 1 & 44 & 10.00\% \\
                                   & bird & 2 & 97 & 22.05\% \\
                                   & drone & 3 & 156 & 35.45\% \\
                                   & helicopter & 4 & 88 & 20.00\% \\
\hline
\textbf{Validation Set (Total: 200)} & airplane & 0 & 25 & 12.50\% \\
                                     & background & 1 & 20 & 10.00\% \\
                                     & bird & 2 & 44 & 22.00\% \\
                                     & drone & 3 & 71 & 35.50\% \\
                                     & helicopter & 4 & 40 & 20.00\% \\
\hline
\textbf{Test Set (Total: 160)}    & airplane & 0 & 20 & 12.50\% \\
                                   & background & 1 & 16 & 10.00\% \\
                                   & bird & 2 & 35 & 21.88\% \\
                                   & drone & 3 & 57 & 35.62\% \\
                                   & helicopter & 4 & 32 & 20.00\% \\
\bottomrule
\end{tabular}
}
\vspace{1mm}
\footnotesize
\end{table}

Model training utilized the cross entropy loss criterion and the \textit{AdamW} optimizer with a learning rate of $1 \times 10^{-4}$ and a weight decay of $1 \times 10^{-2}$. A \textit{ReduceLROnPlateau} scheduler was employed, monitoring validation accuracy with a patience of 5 epochs and a reduction factor of 0.5. Early stopping was implemented with a patience of 20 epochs to prevent overfitting. The model was trained for up to 100 epochs.

\section{Results}

\begin{figure}[h]
 \centering
 \includegraphics[width=1\linewidth]{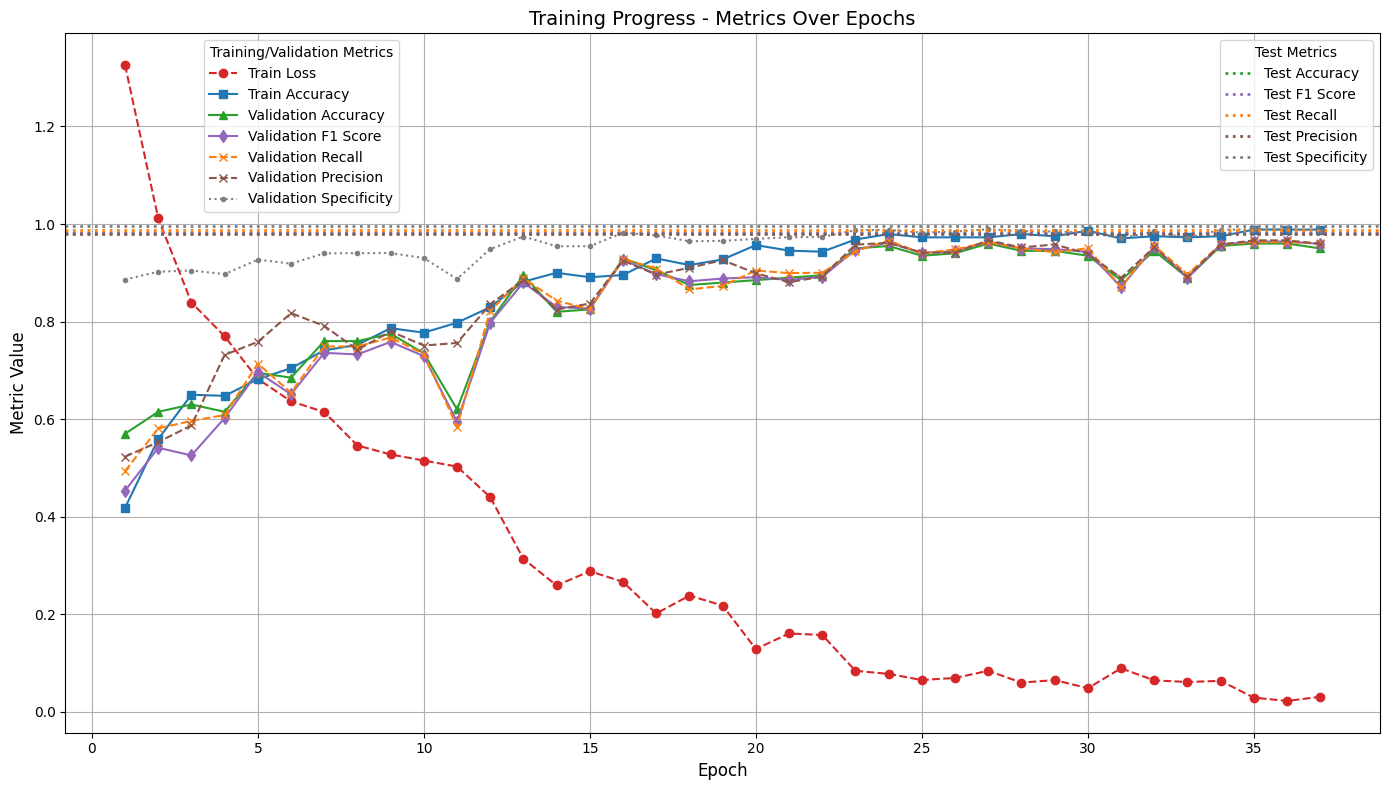}
 \caption{Evolution of Model Performance Metrics: This plot illustrates the evolution of training loss and accuracy, and validation metrics (accuracy, F1-score, recall, precision, specificity) throughout the training epochs. Test set performance is referenced by the horizontal dotted lines. The observed convergence between validation curves and test values, combined with the reduction in training loss, demonstrates the model's generalization ability and absence of overfitting.}
 \label{fig:trainValTest}
\end{figure}

The evaluation of the proposed multimodal Transformer model's performance was rigorously conducted through a comprehensive training, validation, and testing regimen. The entire process, from data loading to the final presentation of results, was completed efficiently. The model's performance was assessed using a suite of metrics: accuracy, macro-averaged recall, macro-averaged precision, macro-averaged F1-score, and macro-averaged specificity. These metrics were calculated from the confusion matrix to provide a comprehensive assessment of the model's ability to classify each class, particularly in the presence of potential class imbalance.

During training, the model demonstrated robust learning and convergence, as depicted in Figure \ref{fig:trainValTest}. The training process culminated at epoch 37 due to the early stopping mechanism. At this point, the average training loss reached 0.0307, with a training accuracy of 0.9886. The validation metrics at the epoch of stopping were: Accuracy 0.9500, Recall 0.9606, Precision 0.9594, F1-score 0.9593, and Specificity 0.9861. The model exhibited a best validation accuracy of 0.9600, and training ceased after 10 epochs without further improvement.

Following the training phase, the model's generalization capability was assessed on an independent test set. The performance metrics on the test set were exceptionally high: Test Accuracy of 0.9812, Test Recall (Macro Avg) of 0.9873, Test Precision (Macro Avg) of 0.9787, Test F1 Score (Macro Avg) of 0.9826, and Test Specificity (Macro Avg) of 0.9954. These results underscore the model's strong ability to generalize to unseen data, indicating effective learning and minimal overfitting.

The computational efficiency of the trained multimodal Transformer model was also rigorously assessed. The model demonstrated a giga floating point operations per second (GFLOPs) count of 1.09 GFLOPs, and it comprises 1.22 million parameters (M). Furthermore, its inference speed on the test set was measured at 41.11 frames per second (FPS). These metrics offer valuable insights into the model's scalability and its potential for real-time applications, highlighting an effective balance between superior predictive performance and computational efficiency.

The model demonstrated robust performance, achieving a Test Accuracy of $0.9812$, with a macro-averaged recall of $0.9873$, a macro-averaged precision of $0.9787$, and a macro-averaged F1-Score of $0.9826$. The high macro-averaged specificity of $0.9954$ further indicates the model's strong ability to correctly identify negative classes. 

\begin{figure}[htb]
 \centering
 \includegraphics[width=0.7\linewidth]{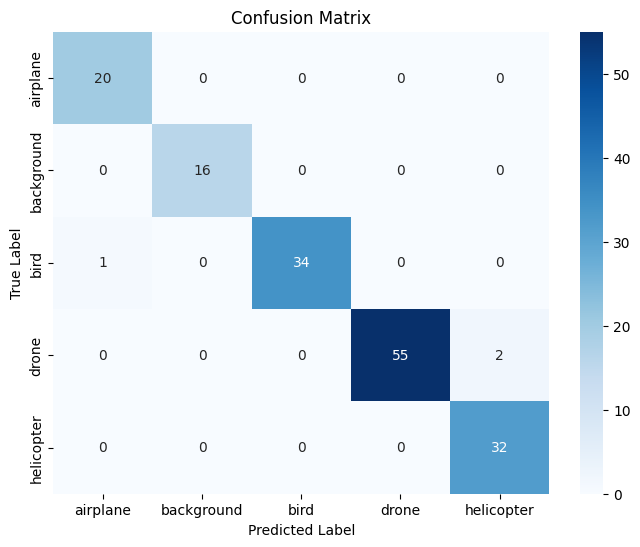}
 \caption{Confusion Matrix highlighting high classification precision and minimal errors across all five classes.}
 \label{fig:confusionMatrix}
\end{figure}

A comprehensive summary of the model's evaluation results, including performance metrics and computational efficiency indicators, is presented in Table~\ref{tab:model_eval}. The confusion matrix in Figure \ref{fig:confusionMatrix} visually details classification performance across all five classes, highlighting true positives, false positives, and false negatives. Analysis reveals high precision for 'airplane', 'background', 'bird', 'drone', and 'helicopter' classes, with most samples correctly identified (55 for 'drone', 34 for 'bird', 20 for 'airplane', 16 for 'background', and 32 for 'helicopter'). Only one misclassification occurred from 'bird' to 'airplane', and two from 'drone' to 'helicopter', demonstrating the model's overall effectiveness and low error rate.

\begin{table}[htbp]
\caption{Performance metrics and computational cost of the final model on the test set.}
\label{tab:model_eval}
\begin{minipage}{\columnwidth}
\begin{center}
\begin{tabular}{llcc}
\toprule
\textbf{Category} & \textbf{Metric} & \textbf{Value} & \textbf{Unit} \\
\hline
Model Evaluation & Test Accuracy & $0.9812$ & - \\
                 & Test Recall (Macro Avg) & $0.9873$ & - \\
                 & Test Precision (Macro Avg) & $0.9787$ & - \\
                 & Test F1 Score (Macro Avg) & $0.9826$ & - \\
                 & Test Specificity (Macro Avg) & $0.9954$ & - \\
\hline
Computational Cost & Floating point operation & $1.09$ & GFLOPs \\
                   & Model parameters & $1.22$ & M \\
                   & Inference speed & $41.11$ & FPS \\
\bottomrule
\end{tabular}
\end{center}
\bigskip
\footnotesize
\end{minipage}
\end{table}

\section{Discussion}

The comprehensive evaluation of the multimodal Transformer model reveals an exceptional level of performance, underscoring its robust capability for aerial object detection and classification. The obtained macro-averaged metrics on the test set—Accuracy of 0.9812, Recall of 0.9873, Precision of 0.9787, F1-score of 0.9826, and Specificity of 0.9954—collectively affirm the model's superior predictive power and generalization ability across diverse aerial object classes.

Each metric provides crucial insights into the practical efficacy of the system:
\begin{itemize}
\item \textbf{Accuracy (0.9812)}: This indicates that approximately 98.12\% of all predictions made by the model are correct. Practically, this translates to an exceedingly reliable system for general classification tasks, minimizing overall misidentification in diverse scenarios.
\item \textbf{Recall (0.9873 Macro Avg)}: A high recall value signifies the model's outstanding ability to correctly identify nearly all actual instances of each class. In practical terms, for critical applications, this means the system has a very low rate of false negatives (missed detections). For example, if 100 drones were present, the model would detect approximately 99 of them, which is vital where the cost of missing an object is high.
\item \textbf{Precision (0.9787 Macro Avg)}: High precision indicates that when the model predicts an object belongs to a certain class, that prediction is highly reliable. This minimizes false positives, meaning fewer false alarms or incorrect interventions. In a security context, this reduces the likelihood of mistaking a non-threat for a threat, thereby optimizing resource allocation and response.
\item \textbf{F1-Score (0.9826 Macro Avg)}: As the harmonic mean of precision and recall, the F1-score provides a balanced assessment of the model's performance. The high F1-score demonstrates that the system achieves both high precision and high recall simultaneously, ensuring that it is robust in both identifying positive cases and ensuring those identifications are correct, even in cases of potential class imbalance.
\item \textbf{Specificity (0.9954 Macro Avg)}: This metric highlights the model's exceptional ability to correctly identify negative cases, i.e., instances that do not belong to a particular class. Practically, this translates to an extremely low false positive rate across all negative instances. For instance, if the system is designed to detect drones, and there are many non-drone objects, it will correctly identify nearly 99.5\% of these as not being drones, significantly reducing irrelevant alerts and improving operational efficiency.
\end{itemize}

A granular analysis of the confusion matrix (Figure \ref{fig:confusionMatrix}) further corroborates these macro-averaged results and provides practical insights into per-class performance. The model demonstrates particularly strong performance in classifying drones, achieving perfect precision (no non-drones were misclassified as drones) and near-perfect recall (only 2 out of 57 drones were misclassified as helicopters). This capability is paramount for security and monitoring applications. Similarly, the background and bird classes exhibit remarkably high precision and recall, with very few misclassifications. The detection of birds, often a challenging task due to their small size and varied appearance, is notably robust.

While the overall performance is outstanding, the confusion matrix identifies a few specific, minor misclassifications that serve as valuable edge cases for future refinement. For instance, one bird was misclassified as an airplane, and two drones were misclassified as helicopters. These isolated instances of inter-class confusion likely stem from inherent visual or spectral similarities between these aerial objects or specific environmental conditions captured in the data. Given the high overall metrics, these are not indicative of systemic flaws but rather nuanced challenges at the boundaries of feature discriminability.

The superior performance metrics achieved by the system are largely attributable to the multimodal Transformer architecture. Its ability to effectively integrate and learn from diverse sensor inputs (audio, RGB and IR video, and radar) is crucial. Unlike unimodal systems, this architecture leverages complementary information, allowing for a more comprehensive understanding of the environment and targets. For example, radar data provides precise range and velocity, while visual data offers rich contextual and identification details, and audio can capture distinct acoustic signatures. The Transformer's self-attention mechanism enables the model to weigh the importance of features across different modalities and time steps, leading to highly robust and discriminative representations. This intrinsic capability to fuse heterogeneous data streams contributes significantly to the model's high recall in detecting objects and its high specificity in accurately categorizing them, making it a highly effective solution for real-world aerial object detection and classification.

\section{Conclusion}

This study successfully developed and rigorously evaluated a novel multimodal Transformer model specifically designed for the classification of aerial objects, integrating heterogeneous data streams from radar, video (RGB and IR), and audio sensors. The research aimed to enhance the precision and robustness of aerial object detection, with a particular focus on distinguishing drones from other airborne entities.

The model achieved exceptional performance on an independent test set, evidenced by macro-averaged metrics of 0.9812 accuracy, 0.9873 recall, 0.9787 precision, 0.9826 F1-score, and 0.9954 specificity. These results comprehensively affirm the model's superior predictive power, robust generalization capabilities, and high reliability across all defined classes. The remarkable recall (0.9873) demonstrates the system's high recall, ensuring that very few actual objects are missed—a critical factor for surveillance and security applications where false negatives can have significant consequences. Concurrently, the high precision (0.9787) indicates the system's strong reliability in its positive identifications, minimizing false alarms and optimizing operational efficiency. This harmonious balance, reflected in the F1-score (0.9826), underscores the model's comprehensive understanding of the multimodal input. Furthermore, the outstanding specificity (0.9954) guarantees that the model is highly effective at correctly identifying non-target instances, preventing misclassifications of innocuous objects.

A key contribution of this research is the demonstrated efficacy of the multimodal Transformer architecture in fusing diverse sensory data. The synergistic combination of radar's precise ranging and velocity information, video's rich contextual and visual identification details, and audio's distinct acoustic signatures, allows the Transformer's self-attention mechanisms to learn intricate intra-modal relationships and, crucially, powerful cross-modal dependencies. This inherent capability to integrate heterogeneous data streams is fundamental to the model's high recall in detecting objects and its high specificity in accurately categorizing them, particularly excelling in the high-precision detection and classification of drones, a paramount capability for modern aerial monitoring systems.

Despite the overall outstanding performance, granular analysis of the confusion matrix revealed a few specific, minor misclassifications, such as isolated instances of birds being misclassified as airplanes, or drones as helicopters. These represent subtle feature overlaps rather than systemic limitations, indicating precise avenues for further refinement.

Future work will focus on several key directions to build upon this robust foundation:
\begin{itemize}
\item Enhanced Feature Engineering: Investigating more advanced feature extraction techniques, particularly for classes exhibiting subtle overlaps, by exploring sophisticated spectral, spatio-temporal, or semantic descriptors that can further improve discriminability.
\item Architectural Optimization: Exploring novel attention mechanisms within the Transformer, optimizing encoder depth, or incorporating explicit temporal modeling techniques to enhance the model's ability to capture long-range dependencies and dynamic object behaviors.
\item Dataset Expansion and Robustness: Expanding the training dataset with greater environmental diversity, challenging conditions (e.g., varying weather, occlusions, sensor noise), and adaptive data augmentation strategies to improve generalization to highly variable real-world operational environments.
\item Real-time Deployment Considerations: Further optimizing computational efficiency for potential edge deployment, including model quantization, pruning, or knowledge distillation, while maintaining high performance.
\end{itemize}

In conclusion, this research presents a significant advancement in multimodal aerial object classification, demonstrating that a well-designed Transformer architecture, leveraging complementary sensor data, can achieve state-of-the-art performance. This work lays a robust foundation for the development of highly accurate, resilient, and intelligent autonomous systems critical for advanced UAV detection in increasingly complex airspaces.

\section{Limitations of the Study}
Despite the outstanding performance achieved by the multimodal Transformer model on the evaluated dataset, a primary limitation of this study pertains to the scope and inherent diversity of the training and test data. The dataset, while meticulously preprocessed and stratified to ensure representative class distribution, was collected under specific environmental conditions and is of a relatively controlled size. Although cyclical replication was employed to augment sample quantity and address potential imbalance in the number of samples per modality, the model's exposure to the full spectrum of real-world operational environments—encompassing varying distances, diverse atmospheric conditions (e.g., fog, rain), complex background clutter, and fluctuating noise levels across different modalities—may be limited. Consequently, while the model demonstrates robust generalization to unseen data within the distribution of the current test set, its performance in significantly different or more challenging real-world scenarios requires further comprehensive validation. This highlights the need for future research to expand the dataset with greater environmental variability to ensure broader applicability and resilience of the proposed multimodal approach in dynamic aerial monitoring contexts.

\printbibliography

\end{document}